\pdfoutput=1

\documentclass[11pt]{article}
\usepackage{acl}
\usepackage{nert}
\usepackage{times}
\usepackage{latexsym}
\usepackage{graphicx}

\usepackage[T1]{fontenc}

\usepackage[utf8]{inputenc}

\usepackage{microtype}

\title{Are UD Treebanks Getting More Consistent? \\A Report Card for English UD}

\author{Amir Zeldes \\
    Georgetown University \\
  \eml{amir.zeldes@georgetown.edu} \\\And
  Nathan Schneider \\
  Georgetown University \\
  \eml{nathan.schneider@georgetown.edu} \\}

\date{}

\begin{document}

\maketitle

\begin{abstract}
Recent efforts to consolidate guidelines and treebanks in the Universal Dependencies project raise the expectation that joint training and dataset comparison is increasingly possible for high-resource languages such as English, which have multiple corpora. Focusing on the two largest UD English treebanks, we examine progress in data consolidation and answer several questions: Are UD English treebanks becoming more internally consistent? Are they becoming more like each other and to what extent? Is joint training a good idea, and if so, since which UD version? Our results indicate that while consolidation has made progress, joint models may still suffer from inconsistencies, which hamper their ability to leverage a larger pool of training data.
\end{abstract}

\section{Introduction}

The Universal Dependencies project\footnote{\url{https://universaldependencies.org}} \cite{10.1162/coli_a_00402} has grown over the past few years to encompass not only over 100 languages, but also over 200 treebanks, meaning several languages now have multiple treebanks with rich morphosyntactic and other annotations. Multiple treebanks are especially common for high resource languages such as English, which currently has data in 9 different repositories, totaling over 762,000 tokens (as of UD v2.11). While this abundance of resources is of course positive, it opens questions about consistency across multiple UD treebanks of the same language, with both theoretical questions about annotation guidelines, and practical ones about the value of joint training on multiple datasets for parsing and other NLP applications.

In this paper we focus on the two largest UD treebanks of English: the English Web Treebank (EWT, \citealt{silveira14gold}) and the Georgetown University Multilayer corpus (GUM, \citealt{Zeldes2017}).\footnote{Due to licensing, GUM Reddit data \cite{behzad-zeldes-2020-cross} has a separate repo, but we merge both repos below.} Although both datasets are meant to follow UD guidelines, their origins are very different: EWT was converted to UD from an older constituent treebank \cite{BiesMottWarnerEtAl2012} into Stanford Dependencies \cite{MarneffeEtAl2006} and then into UD, while GUM was natively annotated in Stanford Dependencies until 2018, then converted to UD \cite{PengZeldes2018}, with more material added subsequently via native UD annotation. Coupled with gradual changes and clarifications to the guidelines, there are reasons to expect systematic dataset differences, which UD maintainers (including the authors) have sought to consolidate from UD version to version.

Despite potential pitfalls, NLP tools are increasingly merging UD datasets for joint training: for example, Stanford's popular Stanza toolkit \cite{QiEtAl2020} defaults to using a model called \texttt{combined} for English tagging and parsing, which is trained on EWT and GUM (including the Reddit subset of GUM).\footnote{Though we focus on English here, the same is true for other UD languages with multiple datasets.} We therefore consider it timely to ask whether even the largest, most actively developed UD treebanks for English are actually compatible; if not, to what extent, and are they inching closer together or drifting apart from version to version? Regardless of the answer to these questions, is it a good idea to train jointly on EWT and GUM, and if so, given constant revisions to the data, since what UD version?

\begin{figure*}[b!t]

\subfloat[EWT]{%
    \label{fig:changes-ewt}
  \includegraphics[clip,width=0.49\textwidth,trim = 0.5cm 0cm 0.5cm 0cm]{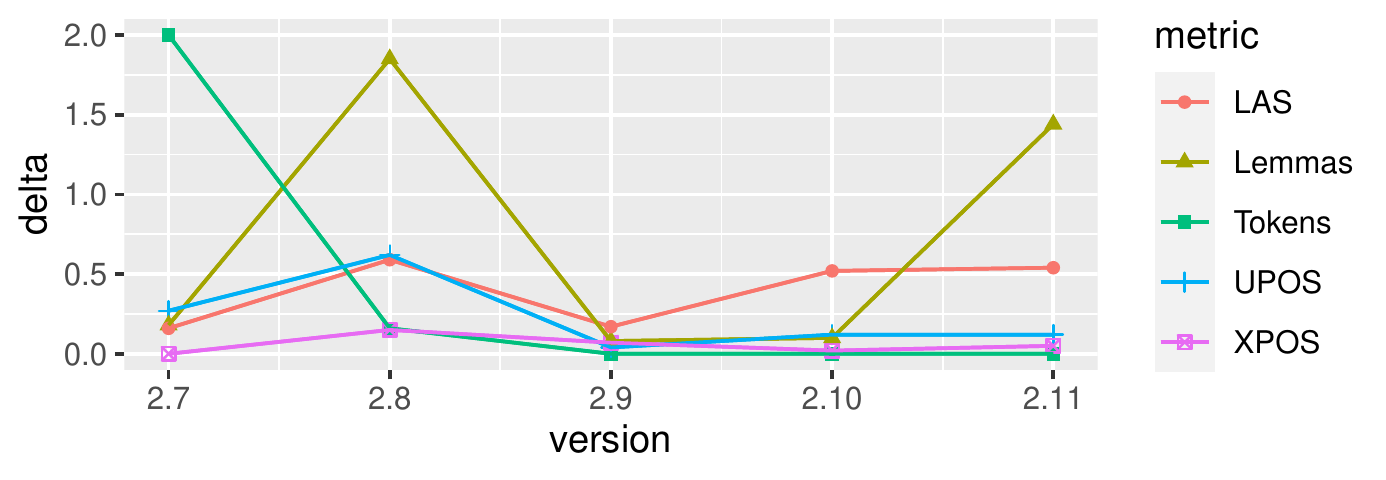}%
}~~\subfloat[GUM]{%
  \includegraphics[clip,width=0.49\textwidth,trim = 0.5cm 0cm 0.5cm 0cm]{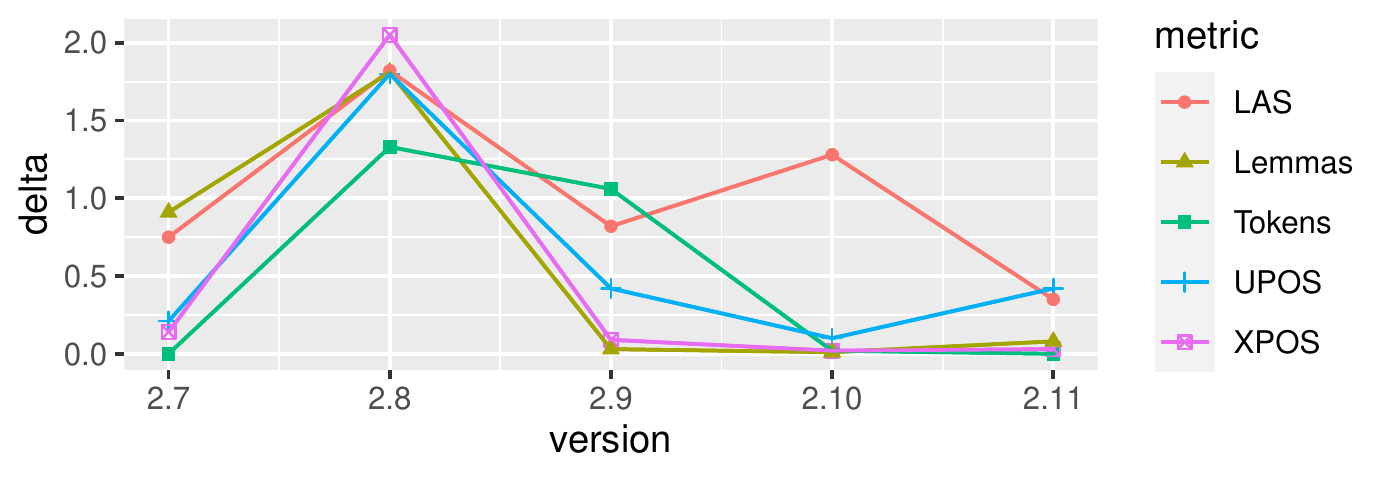}%
}

\caption{Version-to-version changes across annotation layers in EWT and GUM. Y-values are percentages.}\label{fig:changes}
\vspace{-10pt}
\end{figure*}

\section{Related work}

Much previous work on consistency in UD has focused on cross-linguistic comparison, and especially on finding likely errors. Some papers have taken a `breadth-first' automatic approach to identifying any inconsistencies \cite{de-marneffe-etal-2017-assessing}, with the caveat that many types of differences are hard to detect. Others have taken a more focused approach to particular phenomena, for example \citet{bouma-etal-2018-expletives} showed that the \texttt{expl} relation was used differently across languages for comparable cases, using UD v2.1. \citet{SanguinettiEtAl2022} show a broad range of practices in annotating user-generated content from the Web across UD languages in v2.6. \citet{donicke-etal-2020-identifying} also showed inconsistencies within UD languages using UD v2.5, including the finding that two of the top 20 most inconsistently headed relations in UD came from English, where across 7 datasets, \texttt{compound} and \texttt{csubj} behaved differently (of these, only the latter differed substantially in EWT and GUM, though the authors write it is possible that GUM `simply contains more sentences with expletives'). \citet{aggarwal-zeman-2020-estimating} examined part of speech (POS) tag consistency in UD v2.5 and found that POS was relatively internally consistent within most languages. 

Fewer studies have examined cross-corpus parsing accuracy (\citealt{AlonsoZeman2016} for Spanish on UD v1.3, \citealt{DroganovaEtAl2018} for Russian using UD v2.2), and fewer still have looked at parsing consistency and stability \cite{kalpakchi-boye-2021-minor}. However to the best of our knowledge, no previous study has examined changes in consistency across UD versions, i.e.~whether cross-treebank compatibility is increasing over time, how much so, and for which annotations?

\section{How has the data changed?}

To see how data in both corpora has changed across versions, we use the official CoNLL 2018 UD Shared Task \cite{zeman-etal-2018-conll} scorer and compare each of the past six versions of each corpus to its next version, taking the updated version as an improved `gold' standard.\footnote{Earlier comparisons are impossible since they predate GUM's conversion to UD.} This results in a score for each UD metric, such as the labeled attachment score (LAS), universal POS (UPOS) and English-specific POS (XPOS), as well as lemmatization and tokenization. \Cref{fig:changes} shows the difference in score between each pair of versions for each dataset, which we discuss for each corpus below. For example, taking v2.7 of EWT as the correction for v2.6, we see a 2\% rate of tokenization errors (in green), indicating a substantial change in tokenization, but less than 0.2\% change to v2.8, and zero changes to tokenization moving to v2.9.

One caveat to note when working with data across versions is that unlike EWT, GUM's contents are not frozen: the corpus grows with new material every year. In the overview below, we therefore keep the evaluation fixed and limited only to documents that have existed since v2.6 (136 documents, 120K tokens). In \cref{sec:parsing} we will consider scenarios using both this fixed subset and the entire corpus (193 documents, 180K tokens in v2.11).\footnote{The subsequent release of the larger GUM v9, with 203K tokens and 213 documents, was around the same time as the camera-ready deadline for this paper, and could not be evaluated in time.}

\subsection{EWT}

Below we explain the main causes of the larger differentials between consecutive versions.

\paragraph{Tokenization} Multiword tokens (MWTs) were added for most clitics (e.g.~\textit{'ll}) and contractions (\textit{don't}) in v2.7, with some stragglers in v2.8. Essentially no changes to tokenization were made in subsequent versions.

\paragraph{Tagging} Moderate UPOS changes occurred in 2.7 (many WH-words changed to \texttt{SCONJ}) and 2.8 (\texttt{ADJ} and \texttt{VERB} for adjectives and verbs in proper names, formerly \texttt{PROPN}, paralleling the XPOS \texttt{NNP}); this change was followed by GUM as well, see below. XPOS changes were small, peaking in 2.8 for select expressions like \textit{of course}, \textit{at least}, and \textit{United States}.

\paragraph{Lemmatization}
Lemma errors were corrected throughout, but principal sources of lemma changes in v2.8 included capitalization of the content word lemmas in proper names, the lemma for the pronoun \textit{I}, and removal of comparative or superlative degree in the lemmas of \textit{better} and \textit{best}. In v2.11, a new policy for possessive pronoun lemmas was enacted to remove a key discrepancy with GUM.

\paragraph{Dependencies} As shown in \cref{fig:changes-ewt}, the largest changes to LAS occurred in versions 2.8, when newly tagged \texttt{ADJ} tokens in names triggered \texttt{amod}; 2.10, where the analysis of the \textit{X, so Y} construction was changed to \texttt{parataxis} (among others); and 2.11, which featured changes to nesting subjects (\texttt{nsubj:outer}), relative constructions, and clefts.

\newcommand{\stdev}[1]{\hspace{.4em}\small{#1}}
\begin{table*}[bt]
\resizebox{\textwidth}{!}{
\begin{tabular}{ll|cc|cc|cc}
 &  & \multicolumn{2}{c|}{\textbf{EWT test}}  & \multicolumn{2}{c|}{\textbf{GUM test}}  & \multicolumn{2}{c}{\textbf{Macro-Avg}}   \\
\textbf{train} & \textbf{version} & \textbf{UAS (sd)} & \textbf{LAS (sd)} & \textbf{UAS (sd)} & \textbf{LAS (sd)} & \textbf{UAS (sd)} & \textbf{LAS (sd)} \\
 \hline
EWT & v2.6 & 92.82 \stdev{0.132} & 90.24 \stdev{0.066} & 87.81 \stdev{0.073} & 83.89 \stdev{0.023} & 90.31 \stdev{0.059} & 87.07 \stdev{0.025} \\
 & v2.7 & 92.84 \stdev{0.037} & 90.25 \stdev{0.173} & 87.87 \stdev{0.088} & 84.19 \stdev{0.074} & 90.35 \stdev{0.062} & 87.22 \stdev{0.114} \\
 & v2.8 & 92.93 \stdev{0.060} & 90.42 \stdev{0.090} & 87.97 \stdev{0.078} & 84.90 \stdev{0.028} & 90.45 \stdev{0.065} & 87.66 \stdev{0.042} \\
 & v2.9 & 92.88 \stdev{0.107} & 90.41 \stdev{0.131} & 87.57 \stdev{0.148} & 84.36 \stdev{0.105} & 90.23 \stdev{0.098} & 87.38 \stdev{0.117} \\
 & v2.10 & 93.06 \stdev{0.082} & 90.70 \stdev{0.158} & 87.81 \stdev{0.084} & 84.72 \stdev{0.138} & 90.44 \stdev{0.082} & 87.71 \stdev{0.088} \\
 & v2.11 & \textbf{93.18} \stdev{0.142} & \textbf{90.90} \stdev{0.139} & \textbf{88.05} \stdev{0.260} & \textbf{84.74} \stdev{0.289} & \textbf{90.62} \stdev{0.196} & \textbf{87.82} \stdev{0.207} \\
 \hline
GUM & v2.6 & 86.53 \stdev{0.357} & 81.78 \stdev{0.397} & 91.37 \stdev{0.201} & 87.90 \stdev{0.141} & 88.95 \stdev{0.187} & 84.84 \stdev{0.209} \\
 & v2.7 & 86.69 \stdev{0.336} & 82.28 \stdev{0.322} & 91.66 \stdev{0.156} & 88.24 \stdev{0.284} & 89.18 \stdev{0.242} & 85.26 \stdev{0.299} \\
 & v2.8 & 87.02 \stdev{0.133} & 82.90 \stdev{0.214} & 91.88 \stdev{0.132} & 88.86 \stdev{0.159} & 89.45 \stdev{0.002} & 85.88 \stdev{0.041} \\
 & v2.9 & 87.42 \stdev{0.143} & 83.43 \stdev{0.025} & 91.88 \stdev{0.300} & 88.78 \stdev{0.281} & 89.65 \stdev{0.219} & 86.11 \stdev{0.140} \\
 & v2.10 & 87.53 \stdev{0.190} & 83.79 \stdev{0.191} & 92.16 \stdev{0.216} & 89.24 \stdev{0.191} & 89.85 \stdev{0.203} & 86.51 \stdev{0.191} \\
 & v2.11 & \textbf{88.23} \stdev{0.198} & \textbf{84.27} \stdev{0.095} & \textbf{92.28} \stdev{0.137} & \textbf{89.48} \stdev{0.224} & \textbf{90.26} \stdev{0.121} & \textbf{86.88} \stdev{0.132} \\
\end{tabular}
}
\caption{Cross-corpus parsing scores (three run averages with standard deviations)}\label{tab:cross}
\vspace{-10pt}
\end{table*}

\subsection{GUM}

Similarly to  EWT, GUM has become more stable across layers, with little change to XPOS or lemmas since v2.9. However earlier versions show several substantial revisions. Many changes are again simply due to error corrections, but some systematic changes include the following.

\textbf{Tokenization} saw major changes in v2.8, with the introduction of MWTs to match EWT changes. Additional major changes in v2.9 resulted from changing word tokenization to match EWT and other recent LDC corpora, which tokenize hyphenated compounds (e.g.~v2.7 has \textit{data-driven} as one token, but v2.8 has three tokens, like EWT).

\textbf{Tagging} shows a similar shift in v2.8 due to introduction of the \texttt{HYPH} tag for hyphens in compounds like `data-driven', but also the removal of special XPOS tags for square brackets (\texttt{-LSB-} and \texttt{-RSB-} for left/right square brackets were collapsed with the round bracket tags \texttt{-LRB-}/\texttt{-RRB-}, again matching EWT). Changes to UPOS, by contrast, are more substantial, primarily due to verbs/adjectives in proper names, as in EWT above. Later changes to UPOS in v2.9 and 2.11 result from re-tagging some pronominal determiners (XPOS \texttt{DT}) as \texttt{DET} and not \texttt{PRON} (\textit{some}, \textit{all}, \textit{both}), and changing WH subordinators from \texttt{SCONJ} to \texttt{ADV} respectively, again in harmony with changes to EWT.

\textbf{Lemmatization} largely reflects the hyphenation change (since e.g.~`data-driven' is no longer a lemma in v2.8) and the change from \texttt{PROPN} to \texttt{VERB} or \texttt{ADJ} in names, since the lemma for `\textit{Glowing}' in `\textit{the Glowing Sea}' was changed from `\textit{Glowing}' (based on being \texttt{PROPN}) to `\textit{Glow}' (as a \texttt{VERB}).

\textbf{Dependencies} Here too, transition to new tokenization and tagging names created changes in v2.8, but we also see a peak in v2.10, primarily due to consolidation of proper name dependencies (changing \texttt{flat} to syntactically transparent analyses), more aggressive identification of ellipsis (with promoted arguments) and \texttt{orphan} relations, and removal of some uses of the \texttt{dep} relation.

\section{Cross-corpus parsing}\label{sec:parsing}

\begin{table*}[tb]
\resizebox{\textwidth}{!}{
\begin{tabular}{ll|cc|cc|cc}
 &  & \multicolumn{2}{c|}{\textbf{EWT test}}  & \multicolumn{2}{c|}{\textbf{GUM test}}  & \multicolumn{2}{c}{\textbf{Macro-Avg}}   \\
\textbf{train} & \textbf{version} & \textbf{UAS (sd)} & \textbf{LAS (sd)} & \textbf{UAS (sd)} & \textbf{LAS (sd)} & \textbf{UAS (sd)} & \textbf{LAS (sd)} \\
 \hline
\textsc{Joint}$_{subset}$ & v2.6 & 92.38 \stdev{0.044} & 89.59 \stdev{0.108} & 90.08 \stdev{0.366} & 86.80 \stdev{0.326} & 91.23 \stdev{0.177} & 88.20 \stdev{0.146} \\
 & v2.7 & 92.31 \stdev{0.078} & 89.61 \stdev{0.072} & 90.15 \stdev{0.311} & 86.96 \stdev{0.360} & 91.23 \stdev{0.122} & 88.29 \stdev{0.148} \\
 & v2.8 & 92.49 \stdev{0.159} & 89.99 \stdev{0.128} & 90.51 \stdev{0.351} & 87.86 \stdev{0.449} & 91.50 \stdev{0.154} & 88.92 \stdev{0.195} \\
 & v2.9 & 92.39 \stdev{0.324} & 89.80 \stdev{0.278} & 90.63 \stdev{0.392} & 87.91 \stdev{0.415} & 91.51 \stdev{0.086} & 88.85 \stdev{0.114} \\
 & v2.10 & 92.62 \stdev{0.034} & 90.24 \stdev{0.058} & 90.51 \stdev{0.418} & 87.86 \stdev{0.381} & 91.56 \stdev{0.192} & 89.05 \stdev{0.163} \\
 & v2.11 & \textbf{92.92} \stdev{0.072} & \textbf{90.58} \stdev{0.052} & \textbf{90.75} \stdev{0.073} & \textbf{87.94} \stdev{0.059} & \textbf{91.83} \stdev{0.064} & \textbf{89.26} \stdev{0.045} \\
 \hline
\textsc{Joint}$_{all}$ & v2.6 & 92.38 \stdev{0.044} & 89.59 \stdev{0.108} & 90.08 \stdev{0.366} & 86.80 \stdev{0.326} & 91.23 \stdev{0.177} & 88.20 \stdev{0.146} \\
 & v2.7 & 92.31 \stdev{0.078} & 89.61 \stdev{0.072} & 90.15 \stdev{0.311} & 86.96 \stdev{0.360} & 91.23 \stdev{0.122} & 88.29 \stdev{0.148} \\
 & v2.8 & 92.07 \stdev{0.277} & 89.55 \stdev{0.312} & 91.26 \stdev{0.267} & 88.72 \stdev{0.247} & 91.66 \stdev{0.077} & 89.14 \stdev{0.066} \\
 & v2.9 & 92.27 \stdev{0.154} & 89.77 \stdev{0.287} & 90.81 \stdev{0.084} & 88.12 \stdev{0.123} & 91.54 \stdev{0.110} & 88.95 \stdev{0.176} \\
 & v2.10 & 92.18 \stdev{0.018} & 89.86 \stdev{0.010} & 91.54 \stdev{0.170} & 88.99 \stdev{0.211} & 91.86 \stdev{0.092} & 89.43 \stdev{0.110} \\
 & v2.11 & \textbf{92.54} \stdev{0.259} & \textbf{90.11} \stdev{0.240} & \textbf{91.71} \stdev{0.426} & \textbf{89.11} \stdev{0.534} & \textbf{92.13} \stdev{0.147} & \textbf{89.61} \stdev{0.181} \\
\end{tabular}
}
\caption{Joint training parsing scores (three run averages with standard deviations)}\label{tab:joint}
\vspace{-12pt}
\end{table*}

\paragraph{Cross-corpus results} To test whether EWT and GUM are becoming more internally and mutually consistent, we train parsers on each version of each corpus, and test them against both corpora. If each corpus is becoming more consistent, we expect higher scores in each version; and if cross-corpus model scores are increasing, we infer that the data is becoming more consistent across corpora. To ensure a fair comparison, we keep training and test data from GUM fixed to those documents that have been available since v2.6.

The results in \cref{tab:cross} show that within-corpus scores are indeed improving slightly with each version (all scores are 3-run averages using Diaparser, \citealt{attardi-etal-2021-biaffine}, a recent transformer-based biaffine dependency parser). Cross-corpus scores are substantially lower, but also improving: in v2.6, EWT in-domain LAS was 90.24, which has improved slightly to 90.9 in v2.11, but scores training on GUM and testing on EWT have gone up from 81.78 to 84.27. In the opposite direction, GUM in-domain scores improved from LAS=87.9 to 89.48, or for a parser trained on EWT, from 83.89 to 84.74. The macro-average of both corpora also shows a steady increase, more so on GUM. In all cases, v2.11 is the best version yet for all metrics. 

However, since the experiments are limited to the smaller subset of UD GUM v2.6 documents, they do not reflect current NLP tools (which train on all documents in the current UD repos), nor do they tell us whether joint training is a good idea.

\paragraph{Joint training results} In this series of experiments we train on both corpora jointly, comparing two scenarios: the \textsc{subset} scenario limits GUM training data to the v2.6 subset, while \textsc{all} uses all available GUM documents for training at each version; for fairness, scores are always limited to documents in the v2.6 test set, which are a subset of all subsequent release test sets.\footnote{Note that no new documents were added to GUM in v2.7, hence scores are identical for \textsc{subset} and \textsc{all} until v2.8.}

\Cref{tab:joint} shows that here too, there is only improvement over time. Using all GUM documents is superior to just the subset on GUM, but actually leads to a slight degradation on EWT, presumably due to the inclusion of more out-of-domain data from the EWT perspective. Nevertheless, \textsc{Joint}{$_{all}$} performance on EWT also improves over time. 

The best in-domain numbers from \Cref{tab:cross} are always better than the best joint training numbers, indicating that the added data cannot quite compensate for the distraction of different genres in each corpus, and possible remaining annotation inconsistencies. This is not surprising given the importance of genre for NLP performance \cite{zeldes-simonson-2016-different,muller-eberstein-etal-2021-genre}. In fact, similar tradeoffs of a helpful increase in data size vs. a harmful increase in heterogeneity have been observed for UD parsing in other languages (see \citealt{ZeldesEtAl2022} for Hebrew, \citealt{seplnLeon20} for Spanish) and similarly for other tasks (e.g.~for discourse parsing, \citealt{peng-etal-2022-gcdt,LiuZeldes2023}). 

However, the gap is narrowing: the joint model has gained about a point on EWT, placing it only 0.32 points behind the best in-domain model, and it has gained 2.31 points on GUM for the best \textsc{all} scenario in v2.11. Perhaps more importantly, the macroaverage, which may better reflect `real-world' applicability of the parser model to any unseen genre data (since the macro-test set contains the most target genres), is now at LAS=89.61, within one point of the best models for each corpus.

Since the best joint result is also for v2.11, it seems fair to answer the questions posed at the beginning of this paper as follows: it has never been a better idea to train jointly than now; joint training always lags closely behind in-domain training, but the gap has been narrowing and is now very small; and for totally unseen new data, the joint model now looks like a very good idea. The joint \textsc{subset} model is a close second on EWT, and the joint \textsc{all} model is the runner-up on GUM. That said, the fact that more data in the form of a second corpus does not outperform in-domain training alone suggests that there are still inconsistencies between the corpora, on which the next UD versions can hopefully improve.

In terms of concerns about what current jointly trained parsers are actually getting wrong, we direct readers to the confusion matrices in \Cref{fig:confmat} in the Appendix, which indicates that despite training on distinct datasets, the most common errors on both test sets are invariably confusing \texttt{nmod} and \texttt{obl}, which usually corresponds to a PP attachment error. Other systematic errors are rare, and largely concern notable subcategories of names and other types of terms. One recurring subtype is GUM's \texttt{dep} label being confused with EWT \texttt{nummod} for numeric modifiers which are not count-modifiers (e.g.~`Page \textbf{3}' has `3' as \texttt{dep} in GUM but \texttt{nummod} in EWT; GUM only uses \texttt{nummod} for counting cases like `3 pages'). Several of these discrepancies are discussed in \citet{schneider-zeldes-2021-mischievous} and form a target for further consolidation. 

Also of possible concern are \texttt{compound} relations, which a GUM-trained model predicts for various gold-standard relations in EWT, and an EWT model predicts for various gold-standard relations in GUM. It seems likely that these are remaining artifacts from the automatic conversion of the EWT gold constituent annotations to dependencies, in which various complex nominals were analyzed as compounds, for example for names such as \textit{Sri Lanka} or \textit{Hong Kong} (right-headed \texttt{compound} in EWT, but left-headed \texttt{flat} in GUM)
and borrowed foreign words or phrases such as \textit{cordon-blu} (\textit{sic}) (again right-headed in EWT, would be flat in GUM), or also in complex nested phrases which are analyzed as left branching in EWT, e.g.~\textit{Marvel Consultants, Inc.}\ is headed by \textit{Inc.}\ with two \texttt{compound} dependents in EWT. In GUM it would be headed by \textit{Consultants} with \textit{Inc.}\ as \texttt{acl}, or \texttt{flat} for lexicalized cases (attested in GUM for the film \textit{Monsters Inc.}).
Similarly, capitalized adjectival modifiers with XPOS \texttt{NNP} are sometimes labeled as \texttt{compound} in EWT, leading to \texttt{amod} predictions in the GUM-trained model and vice versa (e.g.~\textit{Islamist officers} or \textit{Baathist saboteurs}).


\section{Discussion}

In this paper we surveyed progress in consolidating the largest UD English corpora, EWT and GUM. Results show data is moving closer together: single-corpus training still beats joint training by a hair, but joint models are nearly as good, and likely much more robust. As consolidation continues, we hope to see joint models overtake in-domain training, and more consistency expanding to other English datasets and other UD languages.



\bibliography{anthology,en-ud-progress}

\begin{thebibliography}{25}
\expandafter\ifx\csname natexlab\endcsname\relax\def\natexlab#1{#1}\fi

\bibitem[{Aggarwal and Zeman(2020)}]{aggarwal-zeman-2020-estimating}
Akshay Aggarwal and Daniel Zeman. 2020.
\newblock \href {https://doi.org/10.18653/v1/2020.tlt-1.9} {Estimating {POS}
  annotation consistency of different treebanks in a language}.
\newblock In \emph{Proceedings of the 19th International Workshop on Treebanks
  and Linguistic Theories}, pages 93--110, D{\"u}sseldorf, Germany. Association
  for Computational Linguistics.

\bibitem[{Alonso and Zeman(2016)}]{AlonsoZeman2016}
H\'{e}ctor~Mart\'{i}nez Alonso and Daniel Zeman. 2016.
\newblock \href
  {http://journal.sepln.org/sepln/ojs/ojs/index.php/pln/article/view/5341}
  {Universal {D}ependencies for the {AnCora} treebanks}.
\newblock \emph{Procesamiento del Lenguaje Natural}, 57:91--98.

\bibitem[{Attardi et~al.(2021)Attardi, Sartiano, and
  Simi}]{attardi-etal-2021-biaffine}
Giuseppe Attardi, Daniele Sartiano, and Maria Simi. 2021.
\newblock \href {https://doi.org/10.18653/v1/2021.iwpt-1.19} {Biaffine
  dependency and semantic graph parsing for {E}nhanced {U}niversal
  dependencies}.
\newblock In \emph{Proceedings of the 17th International Conference on Parsing
  Technologies and the IWPT 2021 Shared Task on Parsing into Enhanced Universal
  Dependencies (IWPT 2021)}, pages 184--188, Online. Association for
  Computational Linguistics.

\bibitem[{Behzad and Zeldes(2020)}]{behzad-zeldes-2020-cross}
Shabnam Behzad and Amir Zeldes. 2020.
\newblock \href {https://www.aclweb.org/anthology/2020.wac-1.7} {A cross-genre
  ensemble approach to robust {R}eddit part of speech tagging}.
\newblock In \emph{Proceedings of the 12th Web as Corpus Workshop}, pages
  50--56, Marseille, France. European Language Resources Association.

\bibitem[{Bies et~al.(2012)Bies, Mott, Warner, and
  Kulick}]{BiesMottWarnerEtAl2012}
Ann Bies, Justin Mott, Colin Warner, and Seth Kulick. 2012.
\newblock \href
  {http://www.ldc.upenn.edu/Catalog/catalogEntry.jsp?catalogId=LDC2012T13}
  {English {W}eb {T}reebank}.
\newblock Technical Report {LDC2012T13}, Linguistic Data Consortium,
  Philadelphia, PA.

\bibitem[{Bouma et~al.(2018)Bouma, Hajic, Haug, Nivre, Solberg, and
  {\O}vrelid}]{bouma-etal-2018-expletives}
Gosse Bouma, Jan Hajic, Dag Haug, Joakim Nivre, Per~Erik Solberg, and Lilja
  {\O}vrelid. 2018.
\newblock \href {https://doi.org/10.18653/v1/W18-6003} {Expletives in
  {U}niversal {D}ependency treebanks}.
\newblock In \emph{Proceedings of the Second Workshop on Universal Dependencies
  ({UDW} 2018)}, pages 18--26, Brussels, Belgium. Association for Computational
  Linguistics.

\bibitem[{de~Marneffe et~al.(2017)de~Marneffe, Grioni, Kanerva, and
  Ginter}]{de-marneffe-etal-2017-assessing}
Marie-Catherine de~Marneffe, Matias Grioni, Jenna Kanerva, and Filip Ginter.
  2017.
\newblock \href {https://www.aclweb.org/anthology/W17-6514} {Assessing the
  annotation consistency of the {U}niversal {D}ependencies corpora}.
\newblock In \emph{Proceedings of the Fourth International Conference on
  Dependency Linguistics (Depling 2017)}, pages 108--115, Pisa,Italy.
  Link{\"o}ping University Electronic Press.

\bibitem[{de~Marneffe et~al.(2006)de~Marneffe, MacCartney, and
  Manning}]{MarneffeEtAl2006}
Marie-Catherine de~Marneffe, Bill MacCartney, and Christopher~D. Manning. 2006.
\newblock Generating typed dependency parses from phrase structure parses.
\newblock In \emph{Proceedings of LREC 2006}, pages 449--454, Genoa, Italy.

\bibitem[{de~Marneffe et~al.(2021)de~Marneffe, Manning, Nivre, and
  Zeman}]{10.1162/coli_a_00402}
Marie-Catherine de~Marneffe, Christopher~D. Manning, Joakim Nivre, and Daniel
  Zeman. 2021.
\newblock \href {https://doi.org/10.1162/coli_a_00402} {{Universal
  Dependencies}}.
\newblock \emph{Computational Linguistics}, 47(2):255--308.

\bibitem[{D{\"o}nicke et~al.(2020)D{\"o}nicke, Yu, and
  Kuhn}]{donicke-etal-2020-identifying}
Tillmann D{\"o}nicke, Xiang Yu, and Jonas Kuhn. 2020.
\newblock \href {https://www.aclweb.org/anthology/2020.udw-1.8} {Identifying
  and handling cross-treebank inconsistencies in {UD}: A pilot study}.
\newblock In \emph{Proceedings of the Fourth Workshop on Universal Dependencies
  (UDW 2020)}, pages 67--75, Barcelona, Spain (Online). Association for
  Computational Linguistics.

\bibitem[{Droganova et~al.(2018)Droganova, Lyashevskaya, and
  Zeman}]{DroganovaEtAl2018}
Kira Droganova, Olga Lyashevskaya, and Daniel Zeman. 2018.
\newblock Data conversion and consistency of monolingual corpora: {R}ussian
  {UD} treebanks.
\newblock In \emph{Proceedings of the 17th International Workshopon Treebanks
  and Linguistic Theories (TLT 2018)}, pages 53--66,, Link\"{o}ping, Sweden.
  Link\"{o}ping University Electronic Press.

\bibitem[{Kalpakchi and Boye(2021)}]{kalpakchi-boye-2021-minor}
Dmytro Kalpakchi and Johan Boye. 2021.
\newblock \href {https://aclanthology.org/2021.udw-1.8} {Minor changes make a
  difference: a case study on the consistency of {UD}-based dependency
  parsers}.
\newblock In \emph{Proceedings of the Fifth Workshop on Universal Dependencies
  (UDW, SyntaxFest 2021)}, pages 96--108, Sofia, Bulgaria. Association for
  Computational Linguistics.

\bibitem[{Le{\'{o}}n(2020)}]{seplnLeon20}
Fernando~S{\'{a}}nchez Le{\'{o}}n. 2020.
\newblock \href {http://ceur-ws.org/Vol-2664/capitel\_paper1.pdf} {Combining
  different parsers and datasets for {CAPITEL} {UD} parsing}.
\newblock In \emph{Proceedings of the Iberian Languages Evaluation Forum
  (IberLEF 2020) co-located with 36th Conference of the Spanish Society for
  Natural Language Processing {(SEPLN} 2020), M{\'{a}}laga, Spain, September
  23th, 2020}, volume 2664 of \emph{{CEUR} Workshop Proceedings}, pages 39--44.
  CEUR-WS.org.

\bibitem[{Liu and Zeldes(2023)}]{LiuZeldes2023}
Yang~Janet Liu and Amir Zeldes. 2023.
\newblock Why can’t discourse parsing generalize? {A} thorough investigation
  of the impact of data diversity.
\newblock In \emph{Proceedings of EACL 2023}, Dubrovnik, Croatia.

\bibitem[{M{\"u}ller-Eberstein et~al.(2021)M{\"u}ller-Eberstein, van~der Goot,
  and Plank}]{muller-eberstein-etal-2021-genre}
Max M{\"u}ller-Eberstein, Rob van~der Goot, and Barbara Plank. 2021.
\newblock \href {https://doi.org/10.18653/v1/2021.emnlp-main.393} {Genre as
  weak supervision for cross-lingual dependency parsing}.
\newblock In \emph{Proceedings of the 2021 Conference on Empirical Methods in
  Natural Language Processing}, pages 4786--4802, Online and Punta Cana,
  Dominican Republic. Association for Computational Linguistics.

\bibitem[{Peng et~al.(2022)Peng, Liu, and Zeldes}]{peng-etal-2022-gcdt}
Siyao Peng, Yang~Janet Liu, and Amir Zeldes. 2022.
\newblock \href {https://aclanthology.org/2022.aacl-short.47} {{GCDT}: A
  {C}hinese {RST} treebank for multigenre and multilingual discourse parsing}.
\newblock In \emph{Proceedings of the 2nd Conference of the Asia-Pacific
  Chapter of the Association for Computational Linguistics and the 12th
  International Joint Conference on Natural Language Processing (Volume 2:
  Short Papers)}, pages 382--391, Online only. Association for Computational
  Linguistics.

\bibitem[{Peng and Zeldes(2018)}]{PengZeldes2018}
Siyao Peng and Amir Zeldes. 2018.
\newblock All roads lead to {UD}: Converting {S}tanford and {P}enn parses to
  {E}nglish {U}niversal {D}ependencies with multilayer annotations.
\newblock In \emph{Proceedings of the Joint Workshop on Linguistic Annotation,
  Multiword Expressions and Constructions (LAW-MWE-CxG-2018)}, pages 167--177,
  Santa Fe, NM.

\bibitem[{Qi et~al.(2020)Qi, Zhang, Zhang, Bolton, and Manning}]{QiEtAl2020}
Peng Qi, Yuhao Zhang, Yuhui Zhang, Jason Bolton, and Christopher~D. Manning.
  2020.
\newblock \href {https://nlp.stanford.edu/pubs/qi2020stanza.pdf} {Stanza: A
  {Python} natural language processing toolkit for many human languages}.
\newblock In \emph{Proceedings of the 58th Annual Meeting of the Association
  for Computational Linguistics: System Demonstrations}, pages 101--108.

\bibitem[{Sanguinetti et~al.(2022)Sanguinetti, Cassidy, Bosco, \"{O}zlem
  \c{C}etino\u{g}lu, Cignarella, Lynn, Rehbein, Ruppenhofer, Seddah, and
  Zeldes}]{SanguinettiEtAl2022}
Manuela Sanguinetti, Lauren Cassidy, Cristina Bosco, \"{O}zlem
  \c{C}etino\u{g}lu, Alessandra~Teresa Cignarella, Teresa Lynn, Ines Rehbein,
  Josef Ruppenhofer, Djam\'{e} Seddah, and Amir Zeldes. 2022.
\newblock \href {https://link.springer.com/article/10.1007/s10579-022-09581-9}
  {Treebanking user-generated content: a {UD} based overview of guidelines,
  corpora and unified recommendations}.
\newblock \emph{Language Resources and Evaluation}.

\bibitem[{Schneider and Zeldes(2021)}]{schneider-zeldes-2021-mischievous}
Nathan Schneider and Amir Zeldes. 2021.
\newblock \href {https://aclanthology.org/2021.udw-1.14} {Mischievous nominal
  constructions in {U}niversal {D}ependencies}.
\newblock In \emph{Proceedings of the Fifth Workshop on Universal Dependencies
  (UDW, SyntaxFest 2021)}, pages 160--172, Sofia, Bulgaria. Association for
  Computational Linguistics.

\bibitem[{Silveira et~al.(2014)Silveira, Dozat, de~Marneffe, Bowman, Connor,
  Bauer, and Manning}]{silveira14gold}
Natalia Silveira, Timothy Dozat, Marie-Catherine de~Marneffe, Samuel Bowman,
  Miriam Connor, John Bauer, and Christopher~D. Manning. 2014.
\newblock A gold standard dependency corpus for {E}nglish.
\newblock In \emph{Proceedings of the Ninth International Conference on
  Language Resources and Evaluation (LREC-2014)}.

\bibitem[{Zeldes(2017)}]{Zeldes2017}
Amir Zeldes. 2017.
\newblock \href {https://doi.org/http://dx.doi.org/10.1007/s10579-016-9343-x}
  {The {GUM} corpus: Creating multilayer resources in the classroom}.
\newblock \emph{Language Resources and Evaluation}, 51(3):581--612.

\bibitem[{Zeldes et~al.(2022)Zeldes, Howell, Ordan, and Moshe}]{ZeldesEtAl2022}
Amir Zeldes, Nick Howell, Noam Ordan, and Yifat~Ben Moshe. 2022.
\newblock A second wave of {UD} {H}ebrew treebanking and cross-domain parsing.
\newblock In \emph{Proceedings of {EMNLP} 2022}, Abu Dhabi, UAE.

\bibitem[{Zeldes and Simonson(2016)}]{zeldes-simonson-2016-different}
Amir Zeldes and Dan Simonson. 2016.
\newblock \href {https://doi.org/10.18653/v1/W16-1709} {Different flavors of
  {GUM}: Evaluating genre and sentence type effects on multilayer corpus
  annotation quality}.
\newblock In \emph{Proceedings of the 10th Linguistic Annotation Workshop held
  in conjunction with {ACL} 2016 ({LAW}-X 2016)}, pages 68--78, Berlin,
  Germany. Association for Computational Linguistics.

\bibitem[{Zeman et~al.(2018)Zeman, Haji{\v{c}}, Popel, Potthast, Straka,
  Ginter, Nivre, and Petrov}]{zeman-etal-2018-conll}
Daniel Zeman, Jan Haji{\v{c}}, Martin Popel, Martin Potthast, Milan Straka,
  Filip Ginter, Joakim Nivre, and Slav Petrov. 2018.
\newblock \href {https://doi.org/10.18653/v1/K18-2001} {{C}o{NLL} 2018 shared
  task: Multilingual parsing from raw text to {U}niversal {D}ependencies}.
\newblock In \emph{Proceedings of the {C}o{NLL} 2018 Shared Task: Multilingual
  Parsing from Raw Text to Universal Dependencies}, pages 1--21, Brussels,
  Belgium. Association for Computational Linguistics.

\end{thebibliography}

\appendix 

\section{Confusion matrices}

\begin{figure*}[htb]

\subfloat[Dependency relation errors for cross-corpus training]{%
  \includegraphics[clip,width=1\textwidth,trim = 2cm 0cm 4cm 1.5cm]{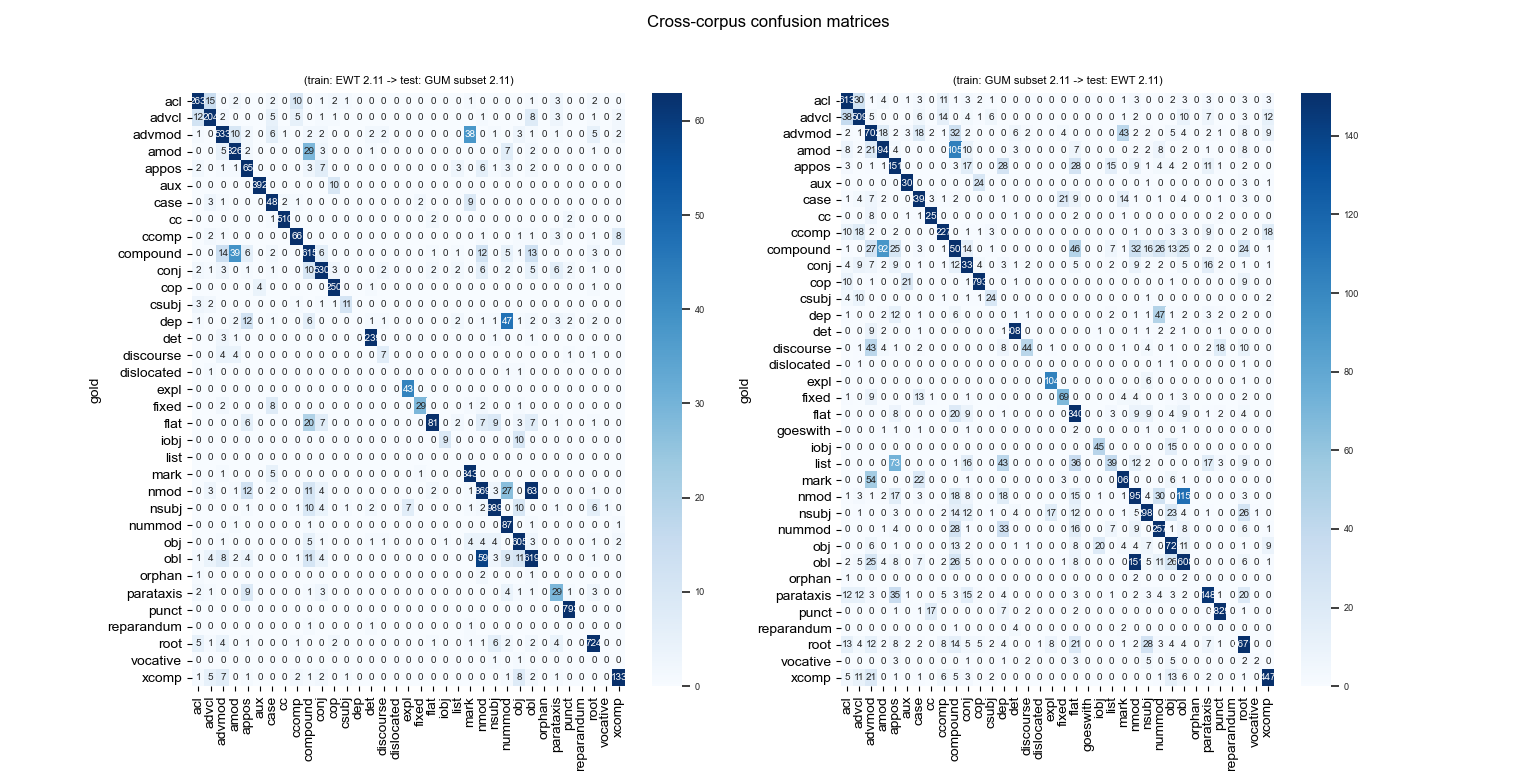}%
}

\subfloat[Dependency relation errors for joint training]{%
  \includegraphics[clip,width=1\textwidth,trim = 2cm 0cm 4cm 1.5cm]{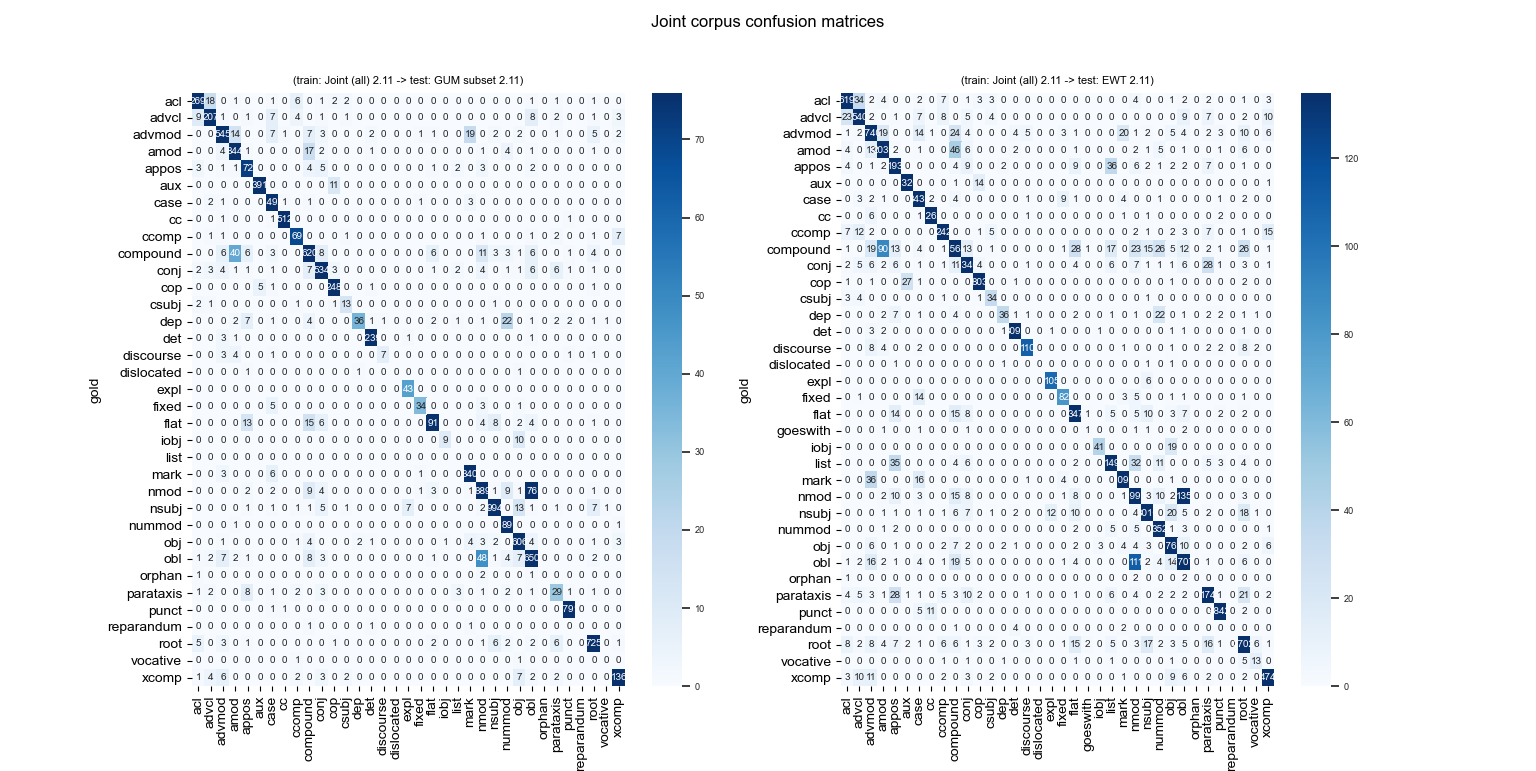}%
}

\caption{Confusion matrices for cross-corpus (a) and joint-corpus (b) dependency relation predictions on both test sets, using the GUM v2.6 document subset for GUM and the average performing parser model from each experiment.}\label{fig:confmat}

\end{figure*}

\Cref{fig:confmat} gives confusion matrices for dependency relation predictions (disregarding correct/incorrect attachment) for the joint and cross-corpus scenarios, testing on GUM (left) and EWT (right). In all cases, the most frequently confused errors are \texttt{obl} and \texttt{nmod} in both directions, largely corresponding to PP attachment ambiguity errors (i.e.~high attachment to the verb for `eat a pizza with a fork' versus low attachment to the object noun in `eat a pizza with anchovies'). These errors are encouraging in that they are unlikely to reflect annotation practice differences between the corpora.

\end{document}